# Combining Multiple-valued Logics in Modular Expert Systems

Jaume Agustí-Cullell    Francesc Esteva    Pere García    Lluís Godó    Carles Sierra

Centre d'Estudis Avançats de Blanes, CSIC
C/ Sta. Barbara s/n
17300 BLANES Girona Spain.
Tel. 34-72-336101 Fax: 34-72-337806
e-mails: agusti,esteva,pere,godo,sierra@ceab.es

## Abstract

The way experts manage uncertainty usually changes depending on the task they are performing. This fact has lead us to consider the problem of communicating modules (task implementations) in a large and structured knowledge based system when modules have different uncertainty calculi. In this paper, the analysis of the communication problem is made assuming that (i) each uncertainty calculus is an inference mechanism defining an entailment relation, and therefore the communication is considered to be inference-preserving, and (ii) we restrict ourselves to the case which the different uncertainty calculi are given by a class of truth-functional Multiple-valued Logics.

## 1 INTRODUCTION

Most expert system (ES) building tools with uncertainty management capabilities provide a unique and global method for representing and combining evidence. Nevertheless, human experts usually change the way they manage uncertainty depending on the task they are performing. To be able to model this behaviour, an ES building tool must allow to attach different uncertainty calculi to the structures implementing the different tasks (in modular ES shells the notion of task is usually implemented as goal-oriented modules). However, tasks or modules in a knowledge base are not independent one of each other, they need to cooperate and communicate, as human experts do when solving complex problems. This can be shown in the following example.

A physician diagnosing a pneumonia could ask to a radiologist about the results of a radiological analysis. The simplest and more frequent type of communication is to get an "atomic" answer like

*"it is likely that the patient has a cavitation in his left lung."*

Then, to use this information in his own reasoning, the physician must only interpret in his language the linguistic expression *likely* used by the radiologist, and perhaps to identify it with another uncertainty term, say for example *acceptable*, used by himself. But the communication could have been richer than that "atomic" answer, and consist of a more complex piece of information. For instance, the radiologist could have answered:

*"if from a clinical point of view you are very confident that the patient has a bacterial disease and he is also inmunodepressed, then its nearly sure he has a cavitation in his left lung."*

As in the previous case, to use the radiologist information the physician must again interpret it. However, this time the interpretation can not be only a matter of uncertainty terms (*very confident, nearly sure*) but also a matter of way of reasoning, if he wants to make use of this information in other situations (i.e., patients) which do not match exactly the one expressed above.

Therefore, if in a knowledge base we have different uncertainty calculi for different tasks (or modules), and these modules need to communicate, a correspondence between their uncertainty calculi must be established. To model the first type of communication shown in the example, in a modular ES shell only a way of translating the languages of different uncertainty calculi, attached to different modules, is required. However, to model the second type of communication the correspondence is also required to be made inference-preserving. The need to preserve sometimes inferences through the communication among tasks can be made clearer by means of another little example from an existing expert system, PNEUMON-IA[1] [Verdaguer, 1989], for the diagnosis of

---

[1] PNEUMON-IA is an application developed in the modular rule-based expert system shell MILORD [Sierra, 1989], that manages linguistically expressed uncertainty (see section 3 for more details).

pneumoniae. The module (task) *Bacteria* of this expert system comprises the following rule:

> *[If a patient has leukopenia and left-deviation then he has a bacterial disease, **sure**]*

stating that the certainty about the bacterianicity of a disease depends on the certainty of the facts *leukopenia* and *left-deviation*, which are investigated in another module named *Laboratory*. Let's suppose that these two modules have different uncertainty calculi. Then we could have two types of communication between them. A first possibility is that *Bacteria* asks to *Laboratory* for the two facts, translates the answers and makes its "and" combination to conclude about bacterianicity. The second possibility is *Bacteria* asks the certainty value of the non-atomic sentence "*leucopenia and left-deviation*" and translates it. In this last case the "and" combination of the certainty values is performed in the *Laboratory* module, and the result is afterwards translated to the *Bacteria* uncertainty calculus and used to conclude about bacterianicity. It seems clear that in order to keep the coherence of the whole diagnosis task, the certainty degree of bacterianicity found out in each case should be the same. To make this sure, the correspondence between the uncertainty calculi of those modules should preserve the inferences made in the *Laboratory* module when moving to the *Bacteria* module.

The general problem of analyzing conditions under which a correspondence or communication between different tasks with different uncertainty calculi preserves inference is a very hard one. In order to deal with this problem, several approaches could be taken into account, from pure logical ones to more cognitive ones. In [Meseguer, 1989], in another but not very different setting, it is argued that "*if the approach taken lacks a logical basis to serve as a criterion of correctness the result may be quite ad hoc and unsatisfactory, and it will probably involve a good deal of costly engineering trial and error*". Following this argument, the approach we have chosen is a logical one, but without forgetting cognitive aspects. More concretely, our analysis will be carried out from two main points:

- first, we will consider uncertainty calculi as inference mechanisms defining logical entailment relationships. Therefore, correspondences (or communications) between different uncertainty calculi will be analyzed as mappings between different entailment systems.

- and second, we will use finite truth-functional multiple-valued logics (MV-logics, for short) as uncertainty calculi, as long as this is a simplified view of the uncertainty reasoning model that our laboratory has been working with in developing applications with the MILORD system [Sierra, 1989], mainly in the medical diagnosis field.

The paper is structured in the following way. After this introductory section, in section two there is a general overview on entailment systems and their inference-preserving mappings. In section three, we describe the class of finite truth-functional multiple-valued logics we will use as uncertainty calculi for different tasks. Section four is devoted to a detailed study of inference-preserving mappings for our uncertainty calculi, and finally, an interactive algorithm for defining such inference preserving correspondences is proposed. This algorithm has been thought as a mechanism to support human experts when developing applications.

## 2 ENTAILMENT SYSTEMS

Inference engines of many rule-based ES can be considered as implementations of proof calculi (from a set of axioms and a set of inference rules) of some underlying Logical Systems. As it is known, every logical system should have a syntactical and semantical formalizations. The theories of Institutions and Entailments Systems allow to formalize an intuitive notion of logical system from the model and proof theoretic approach point of view respectively ([Goguen and Burstall, 1983], [Meseguer, 1989], [Harper et al., 1989]). In this way the Institution approach takes the satisfaction relation between models and sentences as basic whereas the Entailment System approach takes the entailment relation.

The communication problem among tasks or modules has been introduced as the problem of defining inference preserving mappings. Therefore, in this paper we focus our attention in the entailment systems approach rather than in the Institutions one, and thus, we are mainly interested in correspondences between different entailment systems.

Although a categorical definition of Entailment Systems has been given [Meseguer, 1989], for our purposes an *Entailment System* will consist of a pair $(L, \vdash)$, where $L$ is a *language* (a set of sentences, usually built from a set of connectives and a signature that provides a set of sorted symbols), and $\vdash$ is an *entailment relation* on $2^L \times L$, i.e. a relation satisfying the following properties:

E1.- *reflexivity*: for any sentence E, $\{E\} \vdash E$
E2.- *monotonicity*: if $\Gamma \vdash E$ and $\Gamma \subseteq \Gamma'$ then $\Gamma' \vdash E$
E3.- *transitivity*: if $\Gamma \vdash E_i$ for $i \in I$ and
    $\Gamma \cup \{E_i, i \in I\} \vdash E$ then $\Gamma \vdash E$,

where $\Gamma$ and $\Gamma'$ are sets of sentences, and E and $E_i$ are sentences of L.

In [Meseguer, 1989] the following notion of a map of entailment systems has been proposed.

**Definition 2.1. (map of entailment systems)**
*Given the entailment systems $(L, \vdash)$ and $(L', \vdash')$, a mapping $H:L \longrightarrow L'$ is said to be a map of entailment systems if the following condition*
    *If $\Gamma \vdash E$ then $H(\Gamma) \vdash' H(E)$,*
*holds for all set of sentences $\Gamma$ and for all sentence E of L. The map H is said to be conservative if $\Gamma \vdash E$ whenever $H(\Gamma) \vdash' H(E)$.*

A map between entailment systems allows to preserve inference in a strict sense. In particular, when the map is conservative one entailment system is an extension of the other one. However these strong conditions sometimes can be weakened in the uncertainty reasoning framework. From the point of view of the correspondence problem between different tasks with different uncertainty calculi, when a task imports information from another task, it doesn't need always to deduce exactly the same conclusions as the previous one could deduce. Sometimes it only needs that its conclusions be coherent with the deduction system of the other task. In other words, it allows its reasoning to be less accurate when dealing with the other task information, but not incorrect in any case. To model this last situation a definition weaker than the conservative one is introduced below. We will call it *weak conservative*.

**Definition 2.2. (weak conservative map).** *Given entailment systems (L, |-) and (L', |-' ) a map H from L to L' is called weak conservative if the following condition holds:*

*If H(Γ) |-' E' then there exists a sentence E of L
    such that Γ |- E and H(E) |-' E'*

*for all set of sentences Γ of L and all sentence E' of L '.*
*If H is also a map of entailment systems we call it weak conservative map .*

## 3   A CLASS OF MULTIPLE-VALUED LOGICS FOR THE UNCERTAINTY MANAGEMENT IN RULE-BASED EXPERT SYSTEMS

Taking the uncertainty management of MILORD as a reference, in this section we consider a restricted type of MV-logics which are expressive enough to model the uncertainty reasoning used in many rule-based systems. The uncertainty management approach used in MILORD has the following characteristics [Godo et al., 1989]:
1) The expert defines a set of linguistic terms expressing uncertainty corresponding to the verbal scale he will use to weight facts and rules.
2) The set of linguistic terms is supposed to be ordered, at least partially, according to the amount of uncertainty they express, being always the booleans 'true' and 'false' their maximum and minimum elements respectively.
3) The combination and propagation of uncertainty is performed by operators defined over the set of linguistic terms, basically conjunction, disjunction, negation and detachment operators. A method for the elicitation of these operators from the expert has been proposed in (Lopez de Mantaras et al., 90). The main difference of this approach with respect to other ones is that no underlying numerical representation of the linguistic terms is required. Linguistic terms are treated as mere labels. The only a priori requirement is that these labels should represent an ordered set of expressions about uncertainty. For each logical connective, a set of desirable properties of the corresponding operator is listed. Many of these properties are a finite counterpart of those of the uncertainty calculi based on t-norms and t-conorms, which are in turn the basis of the usual [0,1]-valued systems underlying Fuzzy Sets Theory [Alsina et al., 1983]. The listed properties act as constraints on the set of possible solutions. In this way, all operators fulfilling them are generated. This approach has been implemented by formulating it as a constraint satisfaction problem [Godo and Meseguer, 1991]. Finally, the expert may select the one he thinks fits better his own way of uncertainty management in the current task.

These characteristics make clear that the logics associated to the different MILORD uncertainty calculi are a class of finite multiple-valued logics, taking the linguistic terms as truth-values and the operators as the interpretations of the logical connectives. In other words, each linguistic term set, together with its set of operators, defines a truth-values algebra and therefore a corresponding multiple-valued logic. In (Agustí-Cullell et al., 1990), MV-logics have been analyzed from the semantic point of view and formalized as families of Institutions.

Following that line, the main characteristics of our MV-logics for uncertainty management we are concerned with are given by:

- An **algebra of truth-values**: a finite algebra $A = <A_n, 0, 1, N, T, I>$ such that:
   1) The set of truth-values $A_n$ is a chain[2] represented by
   $$0 = a_0 < a_1 < ... < a_{n-1} = 1,$$
   where $0$ and $1$ are the booleans *False* and *True* respectively.
   2) The negation operator N is an unary operation such that the following properties hold:
   N1: if $a < b$ then $N(a) > N(b)$
   N2: $N^2 = Id$.
   3) The "and" operation T is any binary operation such that the following properties hold:
   T1: $T(a,b) = T(b,a)$
   T2: $T(a,T(b,c)) = T(T(a,b),c)$
   T3: $T(0,a) = 0$
   T4: $T(1,a) = a$
   T5: if $a \le b$ then $T(a,c) \not> T(b,c)$ for all c.
   Note that in the unit interval these properties define t-norms functions if we add the condition of continuity.
   4) The implication operator I is defined by residuation with respect to T,.
   $I(a,b) = Max\{ c \in A_n$ such that $T(a,c) \le b \}$
   i.e., I is the finite counterpart of an R-implication generated by the "and" operator T [Trillas and Valverde, 1985].
- A set of **Connectives**: "not"($\neg$), "and"($\&$) and "implication"($\rightarrow$)
- A set of **Sentences**: sentences are pairs of classical-like propositional sentences and intervals of truth-values. The classical like-propositional sentences are built from a set of atomic symbols and the above set of connectives. However the sentences we will consider through this case study are only of the following types:
   $(p_1, V),$
   $(p_1 \& p_2 \& ... \& p_n , V),$
   $(p_1 \& p_2 \& ... \& p_n \rightarrow q, V),$
where $p_1, ..., p_n$ are literals (atoms or negations of atoms), q is an atomic sentence and V is a subset of truth-values. For each truth-values algebra A, $L_A$ will stand for

---

[2] Usually the set of truth-values $A_n$ stands for a totally ordered set of linguistic terms that the expert uses to express uncertainty, but nothing changes if it is only partially ordered.

the set of sentences with intervals of truth-values belonging to A
- **Models:** are defined by valuations, i.e. mappings $\rho$ from to the firsts components of sentences to $A_n$ provided that:

$\rho(\neg p) = N(\rho(p))$
$\rho(p_1 \& p_2) = T(\rho(p_1), \rho(p_2))$
$\rho(p \rightarrow q) = I(\rho(p), \rho(q))$

- **Satisfaction Relation:** between models and sentences is defined by:

$M_\rho \models (p, V)$ if, and only if $\rho(p) \in V$,

where $M_\rho$ stands for the model defined by a valuation $\rho$.
- **Entailment Relation:** the minimal one generated by
   1) the following set of axioms:
      (A-1) $((p_1 \& p_2) \& p_3 \rightarrow p_1 \& (p_2 \& p_3), 1)$
      (A-2) $(p_1 \& (p_2 \& p_3) \rightarrow (p_1 \& p_2) \& p_3, 1)$
      (A-3) $(p_1 \& p_2 \rightarrow p_2 \& p_1, 1)$
      (A-4) $(\neg\neg p \rightarrow p, 1)$
   2) the following inference rules, which are sound with respect to the satisfaction relation (Agustí-Cullell et al., 1990):
      (RI-1) WEAKENING: $\Gamma, (p, V) \vdash (p, V')$,
      where $V \subseteq V'$ and $\Gamma$ is a set of sentences,
      (RI-2) NOT-introduction: $(p, V) \vdash (\neg p, N(V))$,
      (RI-3) AND-introduction:
      $(p_1, V_1), (p_2, V_2) \vdash (p_1 \& p_2, T(V_1, V_2))$,
      (RI-4) MODUS PONENS:
      $(p, V_1), (p \rightarrow q, V_2) \vdash (q, MP(V_1, V_2))$, being

$$MP(a,b) = \begin{cases} \emptyset, & \text{if } a \text{ and } b \text{ are inconsistent} \\ [a,1], & \text{if } b = 1 \\ T(a,b), & \text{otherwise} \end{cases}$$

where $a$ and $b$ are said to be inconsistent if there exists no $c$ such that $I(a,c) = b$.

Notice that these inference rules are the only ones that an inference engine would need when working on sets of sentences of the above specified types, very common in fact in rule-based ES. However, instead of the rule RI-4 and for the sake of simplicity, we will consider the following inference rule:

(RI-4') MODIFIED MODUS PONENS:
$(p, V_1), (p \rightarrow q, V_2) \vdash (q, T(V_1, V_2))$.

Although it is correct for instance in the usual case of upper intervals of truth-values, this inference rule is not logically sound in general with respect to the semantics (satisfaction relation) above defined. Nevertheless, it is a well known fact that, from the cognitive point of view, detachment operators share the same properties required to conjunction operators [Bonissone, 1987]. These arguments, together with self-evident simplicity reasons, have lead us to adopt the inference rule RI-4'. Therefore, from now on, given a truth-values algebra A we will denote by $MVL_A$ the multiple-valued logic defined above, and by $(L_A, \vdash_A)$ its associated entailment system. The language of this entailment system is $L_A$ and its entailment relation is the minimal one determined by axioms A-1, A-2, A-3 and A-4, and by inference rules RI-1, RI-2, RI-3 and RI-4'.

On the other hand, the disjunction operator needed for parallel combination can be obtained from the negation and conjunction operators using the De Morgan laws. For these reasons, and for deductive purposes, only the ordered set of truth-values (linguistic terms) and the conjunction and negation operators should be specified in the truth-values algebra definitions. Therefore, from now on, truth-values algebras will be represented by <A,T,N>, omitting the booleans **0** and **1**, as long as they belong to all algebras.

## 4   INFERENCE PRESERVING MAPS BETWEEN MV-LOGICS

The aim of this section is to analyze the problem of preserving inference in communicating modules, assuming that each one has its own finite MV-logic as uncertainty calculus. In section two, maps and weak conservative maps of entailment systems have been introduced in order to model inference preserving correspondences. In the first subsection of this section, it is shown that morphisms and quasi-morphisms of truth-values algebras generate maps and weak conservative maps respectively of the corresponding entailment systems. In the second and third subsection, morphisms and quasi-morphisms of truth-values algebras are studied. Finally, in the fourth subsection an interactive algorithm to define such mappings to assist human experts when developing applications, is proposed.

### 4.1   WEAK CONSERVATIVE MAPS

Now we consider the problem of finding inference preserving correspondences between two of these logics $MVL_A$ and $MVL_B$, where $A = <A_n, T, N>$ and $B = <B_m, T', N'>$ are their corresponding truth-values algebras. As it has been noted in section 2, the mappings between their entailment systems $(L_A, \vdash_A)$ and $(L_B, \vdash_B)$ we are mainly interested in are the *weak conservative* ones. In order to give a sufficient condition for a mapping $f: A_n \longrightarrow B_m$ to generate a weak conservative mapping between the entailment systems of $MVL_A$ and $MVL_B$, we need some new definitions and results:

Given a truth-values algebra $A = <A_n,T,N>$, we consider the set of intervals of $A_n$, $I(A_n) = \{[a,b] \mid a,b \in A_n\}$ where $[a,b] = \{x \mid x \in A_n, a \leq x \leq b\}$. We can define the following order relation in $I(A_n)$:

*$I_1 \leq^* I_2$ if $a \leq b$ for all $a \in I_1$ and for all $b \in I_2$.*

Let's consider now the following operations on $I(A_n)$
   1) $N^*([a,b]) = [N(b), N(a)]$
   2) $T^*([a_1,b_1],[a_2,b_2]) = [T(a_1,a_2), T(b_1,b_2)]$

It is easy to check that $N^*$ is a negation mapping and $T^*$ fulfils T1÷T5. Moreover, identifying every element $a$ of $A_n$ with the interval $[a,a]$ of $I(A_n)$, $<A_n, T, N>$ is a subalgebra of $<I(A_n),T^*,N^*>$, that is, we have the following proposition.

**Proposition 4.1.** *Any truth-value algebra $<A_n, T, N>$ can be extended to an algebra $<I(A_n), T^*, N^*>$ of the same type that has $<A_n, T, N>$ as a subalgebra.*

It is worth noticing that $(I(A_n), \leq^*)$ is only a partial ordered set with minimum $0 = [0,0]$ and maximum $1 = [1,1]$, and $N^*$ and $T^*$ are univocally defined by $N$ and $T$. Next we give a small example of an algebra of intervals generated by a truth-values algebra of four elements

**Example.** Let $A = \{0 < a < b < 1\}$ the chain of four elements. The set of intervals of A is $I(A) = \{[0,a], [0,b], [0,1], [a,b], [a,1], [b,1], [0,0], [a,a], [b,b], [1,1]\}$. Identifying every interval $[x,x]$ with the element $x$ of A, the order relation on A and I(A) can be represented by the graphs of figure 1.

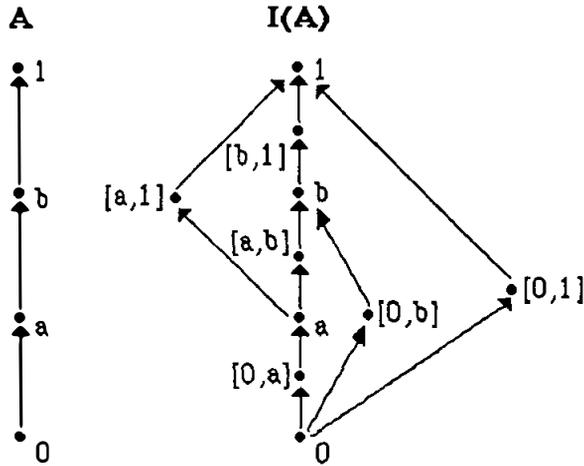

Figure 1: Graphs of the order relation of A and I(A).

Next we introduce what we call a **quasi-morphism** of algebras as a weakened notion of morphism, allowing to map values of an algebra into intervals of the other. This tries to capture the possibility of being imprecise when renaming truth-values from a MV-logic to another.

**Definition 4.1.** *Given two truth-values algebras $A = <A_n, T_1, N_1>$ and $B = <B_m, T_2, N_2>$, a mapping $f: A_n \longrightarrow I(B_m)$ is a quasi-morphism from A to B if the following conditions hold:*
   *1) f is non-decreasing, i.e. if $a \leq b$, then $f(a) \nleq f(b)$,*
   *2) $f(0) = 0$,*
   *3) $f(N_1(x)) = N_2^*(f(x))$,*
   *4) $f(T_1(x,y)) \subset T_2^*(f(x), f(y))$.*

It is clear from this definition that :
   (i) algebra morphisms are a particular case of algebra quasi-morphisms, identifying every element $b$ of B with the interval $[b,b]$. Moreover, a quasi-morphism f is a morphism if, and only if, $f(A_m) \subset B_m$.
   (ii) algebra morphisms from $<A_n, T_1, N_1>$ to $<I(B_m), T_1^*, N_2^*>$ are also quasi-morphisms.

**Theorem 4.2.** *Let $MVL_A$ and $MVL_B$ be the multiple-valued logics defined on the truth-values algebras $A = <A_n, T, N>$ and $B = <B_m, T', N'>$ respectively. Then every quasi-morphism of A to B generates a weak conservative mapping between the entailment systems of $MVL_A$ and $MVL_B$.*

*Proof.* Let $f: A_n \longrightarrow I(B_m)$ be a quasi-morphism between A and B, and let $\Gamma = \{(p_1, V_1), ..., (p_n, V_n)\}$ a set of sentences of $MVL_A$. We will denote by $H_f$ the translation function from $MVL_A$ to $MVL_B$ defined by $H_f((s,V)) = (s,f(V))$. Obviously $H_f$ translates axioms of $MVL_A$ into axioms of $MVL_B$. Suppose that in $MVL_B$ a sentence $E' = (q, V')$ can be derived from $H_f(\Gamma) = \{(p_1, f(V_1)), ..., (p_n, f(V_n))\}$ and axioms of $MVL_B$ by applying a sequence of RI-1, RI-2, RI-3 and RI-4' inference rules and let's denote by $g'$ the composition of their corresponding inference functions (only functions N' and T' will appear). Then it must be the case that $g'(f(V_1), ..., f(V_n)) \subset V'$. We have to show that there exists a sentence E of $MVL_A$ such that it can be derived from $\Gamma \vdash_A E$ and that $H_f(E) \vdash_B E'$. Let E be the sentence $(q, g(V_1, ..., V_n))$, where $g$ is the function obtained from g' replacing the occurrences of N' and T' by N and T. Then it is clear that $\Gamma \vdash_A E$ holds and, because f is a quasi-morphism, we have $f(g(V_1, ..., V_n)) \subset g'(f(V_1), ..., f(V_n)) \subset V'$, so $H_f(E) \vdash_B E'$ also holds and the theorem have been proved.◆

In the particular case of algebra morphisms, the following proposition shows that they also generate maps of entailment systems.

**Proposition 4.3.** *Let $MVL_A$ and $MVL_B$ be the multiple-valued logics defined on the truth-values algebras $A = <A_n, T, N>$ and $B = <B_m, T', N'>$ respectively. Then every order-preserving morphism of A to B generates a map between the entailment systems of $MVL_A$ and $MVL_B$.*

### 4.2 MORPHISMS OF TRUTH-VALUES ALGEBRAS

In this subsection we turn our attention to the problem of relating truth-values algebras by means of order-preserving mappings which are algebra morphisms, and some necessary and/or sufficient conditions for their existence are given. Although there can exist algebra morphisms which are not order-preserving, for cognitive and logical reasons it seems reasonable to require this condition. Therefore, as long as we are only interested in order-preserving mappings, from now on, and for the sake of simplicity, we will use the term *morphisms* as an abbreviation of order-preserving morphism with $f(0) = 0$.

We begin with some well-known results on negation operators. For each chain $A_n$, there exists only one negation N and it is defined by
$$N(a_i) = a_{n-i-1}.$$
Then every chain $A_n$ can be partitioned in three subsets:
- the set of negative elements $\mathbf{N}_n = \{x \mid x < N(x)\}$
- the set of fixed elements $\mathbf{F}_n = \{x \mid x = N(x)\}$
- the set of positive elements $\mathbf{P}_n = \{x \mid x > N(x)\}$

being these subsets $\mathcal{F}_n = \{ a_k \}$, $\mathcal{N}_n = \{ a_i \mid i<k \}$, $\mathcal{P}_n = \{ a_i \mid i>k \}$, if n =2k+1, and $\mathcal{F}_n = \emptyset$, $\mathcal{N}_n = \{ a_i \mid i<k \}$ and $\mathcal{P}_n = \{ a_i \mid i \geq k \}$, if n=2k.

The following equalities also hold: $N(\mathcal{N}_n) = \mathcal{P}_n$, $N(\mathcal{F}_n) = \mathcal{F}_n$ and $N(\mathcal{P}_n) = \mathcal{N}_n$.

**Proposition 4.4.** *Given two chains $A_n$ and $B_m$, a mapping $f: A_n \longrightarrow B_m$ is a morphism with respect to the negation operator if, and only if, the following conditions hold:*

1. $f_{|\mathcal{N}_n}$ *is an order preserving mapping from $\mathcal{N}_n$ to $\mathcal{N}_m \cup \mathcal{F}_m$ such that $f(0) = 0$*
2. $f(\mathcal{F}_n) = \mathcal{F}_m$
3. *If $a_i$ belongs to $\mathcal{P}_n$, $f(a_i) = N'(f(a_{n-i-1}))$, where $N'$ is the negation associated to $B_m$.*

From this proposition it is easy to show that:
- If $n$ is odd then, in order to be f a *morphism*, m must be also odd, as long as $f(\mathcal{F}_n) = \mathcal{F}_m \neq \emptyset$.
- In the case of being $n$ even or both $n$ and $m$ odd, every mapping $f_1: \mathcal{N}_n \longrightarrow \mathcal{N}_m \cup \mathcal{F}_m$ defines a negation morphism $f$ in the following way:

$$f(a_i) = \begin{cases} f_1(a_i), & \text{if } a_i \in \mathcal{N}_n \\ b_r, & \text{if } a_r \in \mathcal{F}_n \text{ and } \mathcal{F}_m = \{b_r\} \\ N'(f_1(N(a_i))), & \text{if } a_i \in \mathcal{P}_n \end{cases}$$

and reciprocally, every negation morphism $f$ is defined by the mapping $f_1 = f_{|\mathcal{N}_n}$.

We follow now with two propositions about how "and" operations can defined in order to have algebra morphisms.

**Proposition 4.5.** *Let $< A_n, N, T >$ be a truth-values algebra and let $B_m$ a chain containing $A_n$ such that the negation $N'$ associated to B fulfils $N'/A_n = N$, i.e. if $a_i = b_j$ then $N(a_i) = b_{m-j-1}$. Then there exists at least an "and" operation $T'$ on $B_m$ such that $<A_n, N, T>$ is a subalgebra of $< B_m, N', T' >$.*

*Proof.* It can be easily checked that the mapping T' defined as:

$$T'(p,q) = \begin{cases} T(p^-, q^-), & \text{if } p \neq 1 \text{ and } q \neq 1 \\ p, & \text{if } q = 1 \\ q, & \text{if } p = 1 \end{cases}$$

where $p^- = \max\{ x \in A_n \mid x \leq p \}$, is an "and" operation on $B_m$, and that it also verifies $T'_{|A_n \times A_n} = T$. ◆

**Proposition 4.6.** *Let $<A_n, T, N>$ be a truth-values algebra and let $f: A_n \longrightarrow B_m$ be a negation morphism. Then there exists an "and" operation $T'$ on $B_m$ such that $f$ is an algebra morphism if, and only if, $f$ is compatible with $T$, that is, for all $a,b,c,d \in A_n$, $f(a) = f(b)$ and $f(c) = f(d)$ imply $f(T(a,c)) = f(T(b,d))$.*

*Proof.* Obviously, if f is an algebra morphism it is compatible with T. On the other hand, if f is compatible with T, the relation $\approx_f$ defined as

$$a \approx_f b \quad \text{if}_{\text{DEF}} \quad f(a) = f(b)$$

is a congruence relation on $A_n$. Let $< A_n/\approx_f, T_f, N_f >$ be the quotient algebra. Therefore, identifying $A_n/\approx_f$ with $f(A_n)$, f will be a morphism from $<A_n, T, N>$ to $< f(A_n), T_f, N_f>$. By proposition 4.5 there exists T' on $B_m$ such that $< f(A_n), T_f, N_f>$ is a subalgebra of $<B_m, T', N'>$, where N' is the negation associated to $B_m$. ◆

Now, taking into account the above results, it is interesting to point out some considerations about the problem of algebra morphism generation. In the following A will stand for a truth-values algebra $<A_n, T, N>$ and $B_m$ for a chain of m elements.

1) If $n$ is odd and $m$ even, there is no possible morphism between A and B, being B any truth-values algebra defined on $B_m$. Then, this case will not be considered any more.

2) In order to define a morphism $f: A_n \longrightarrow B_m$ with respect to the negation operator we only need to take any mapping $f_1: \mathcal{N}_n \longrightarrow \mathcal{N}_m \cup \mathcal{F}_m$ with $f_1(0) = 0$ and to extend it in the way above indicated.

3) The generation of possible operations T' on $B_m$, together with renaming mappings f from $A_n$ to $B_m$ such that f: $<A_n, T, N> \longrightarrow <B_m, T', N'>$ are morphisms, is a process that can be automated without difficulties. The problem reduces to:
- first, to generate all mappings f from all order preserving mappings $f_1: \mathcal{N}_n \longrightarrow \mathcal{N}_m \cup \mathcal{F}_m$ with $f_1(0) = 0$,
- second, to check which ones are compatible with T,
- and third, to generate all algebras $B_m$ containing $f(A_m)$ as subalgebra.

This method will provide us with a family of suitable algebras $<B_m, T', N'>$ for each mapping $f_1$.

4) Nevertheless, in general, it can be the case that the set of possible solutions would be empty. As an example, consider the algebra $<A_4, T, N>$, where $A_4 = \{0 < a_1 < a_2 < 1\}$, and the "and" operation T is the one given in figure 2.

| T | 0 | $a_1$ | $a_2$ | 1 |
|---|---|---|---|---|
| 0 | 0 | 0 | 0 | 0 |
| $a_1$ | 0 | 0 | 0 | $a_1$ |
| $a_2$ | 0 | 0 | $a_1$ | $a_2$ |
| 1 | 0 | $a_1$ | $a_2$ | 1 |

Figure 2: "And" operation in a chain of four elements

It is easy to check that there exists no renaming mapping from $A_4$ to $B_3$ such that a morphism between $<A_4, T, N>$ and $<B_3, T', N'>$ can be defined for any "and" operation T'. Let $B_3 = \{0 < b < 1\}$, and thus $\mathcal{N}' = \{0\}$, $\mathcal{F}' = \{b\}$, and $\mathcal{P}' = \{1\}$. It is clear that if $f_1: \{0, a_1\} \longrightarrow \{0, b\}$ is order-preserving and $f_1(0) = 0$, then there are only two possibilities:

1) $f_1(a_1) = 0$, and then the mapping $f: A_4 \longrightarrow B_3$ is defined by $f(0) = f(a_1)$ and $f(a_2) = f(1) = 1$, but f is not compatible with T, because $f(a_2) = f(1)$ but $f(T(a_2, a_2)) = 0 \neq 1 = f(T(a_2, 1))$;

2) $f_1(a) = b$, and then the mapping $f: A_4 \longrightarrow B_3$ is defined by $f(0) = 0$, $f(a_1) = f(a_2) = b$, and $f(1) = 1$, but again f is not compatible with T, because $f(a_1) = f(a_2)$ but $f(T(a_1,a_1)) = 0 \neq b = f(T(a_2,a_2))$.

## 4.3 QUASI-MORPHISMS OF TRUTH-VALUES ALGEBRAS

As we have seen in last section, given any truth-values algebras A and B, it is not always possible to find a morphism between them. However, this is not the case of quasi-morphisms because of the additional freedom of mapping (or renaming) a truth-value of A to an interval of B. This point is proved in the following proposition.

**Proposition 4.7**: *Let $A = <A_n, T_1, N_1>$ and $B = <B_m, T_2, N_2>$ be two truth-values algebras. Let $C = <C_k, T', N'>$ be an algebra that can be imbedded in both A and B, and let $h_1$ and $h_2$ their corresponding monomorphisms. Then there exists at least one quasi-morphism f from A to B such that $f(h_1(c)) = h_2(c)$, for all $c \in C_k$.*

*Proof.* First of all, notice that given A and B there always exists the algebra C, because at least the algebra of booleans satisfies the required condition for any pair of truth-values algebras. So, let $h_1$ and $h_2$ be the corresponding monomorphisms from C to A and B respectively, and consider the mapping $f: A_n \longrightarrow I(B_m)$ defined by:

$$f(x) = [h_2(c_x^-), h_2(c_x^+)]$$

where $c_x^- = \max\{c \text{ of } C_k \mid h_1(c) \leq x\}$ and $c_x^+ = \min\{c \text{ of } C_k \mid h_1(c) \geq x\}$. Straightforward computation shows that the required properties for f to be a quasi-morphism hold.♦

As an example, let's consider the algebras $A = <A_5, T, N>$ and $B = <B_7, T', N'>$, where the "and" operations T and T' are given in figure 3 respectively.

| T | 0 | $a_1$ | $a_2$ | $a_3$ | 1 |
|---|---|---|---|---|---|
| 0 | 0 | 0 | 0 | 0 | 0 |
| $a_1$ | 0 | 0 | $a_1$ | $a_1$ | $a_1$ |
| $a_2$ | 0 | $a_1$ | $a_2$ | $a_2$ | $a_2$ |
| $a_3$ | 0 | $a_1$ | $a_2$ | $a_3$ | $a_3$ |
| 1 | 0 | $a_1$ | $a_2$ | $a_3$ | 1 |

| T' | 0 | $b_1$ | $b_2$ | $b_3$ | $b_4$ | $b_5$ | 1 |
|---|---|---|---|---|---|---|---|
| 0 | 0 | 0 | 0 | 0 | 0 | 0 | 0 |
| $b_1$ | 0 | $b_1$ | $b_1$ | $b_1$ | $b_1$ | $b_1$ | $b_1$ |
| $b_2$ | 0 | $b_1$ | $b_2$ | $b_2$ | $b_2$ | $b_2$ | $b_2$ |
| $b_3$ | 0 | $b_1$ | $b_2$ | $b_3$ | $b_3$ | $b_3$ | $b_3$ |
| $b_4$ | 0 | $b_1$ | $b_2$ | $b_3$ | $b_4$ | $b_4$ | $b_4$ |
| $b_5$ | 0 | $b_1$ | $b_2$ | $b_3$ | $b_4$ | $b_5$ | $b_5$ |
| 1 | 0 | $b_1$ | $b_2$ | $b_3$ | $b_4$ | $b_5$ | 1 |

Figure 3. "And" operations in chains of five and seven elements

It can be checked that there is no morphism from A to B. On the other hand $C = <C_3, T'', N''>$, being $C_3 = \{0<b<1\}$ and $T''(x,y) = \min(x,y)$, is the maximal subalgebra that can be imbedded into A and B, and the monomorphisms $h_1$ and $h_2$ are defined by:

$h_1(0) = 0$     $h_2(0) = 0$
$h_1(b) = a_2$   $h_2(b) = b_3$
$h_1(1) = 1$     $h_2(1) = 1$

respectively. The above proposition 4.7 assures that the mapping $f: A_5 \to I(B_7)$ defined by:

$f(0) = 0$
$f(a_1) = [0, b_3]$
$f(a_2) = b_3$
$f(a_3) = [b_3, 1]$
$f(1) = 1$

is a quasi-morphism from A to B.

## 4.4 PRAGMATICS OF INFERENCE PRESERVING MAPPINGS

In the previous sections we have seen that when two modules, that use different multiple-valued logical languages, need to communicate, an inference preserving mapping between their logics has to be defined. The process of defining these mappings could be supported by an interactive algorithm based on the notion of morphism and quasi-morphism between truth-values algebras. This process could be as follows:

Initially the expert determines a MV-logic for each module, giving the set of truth-values and the "and" truth-table, and proposes as many as needed renaming functions between different truth-values sets. Then, for each pair of modules having different MV-logics, the interactive algorithm will take their truth-values algebras and the renaming between them and check if the initial renaming is an algebra quasi-morphism. If not, the algorithm presents to the expert the set of possible modifications of the renaming function, each one being a morphism or a quasi-morphism. The expert may then select the one that fits better his aims. If no selection is made, possible modifications of the "and" truth-table are suggested, keeping the initial renaming function. Finally, if there has not been any solution, possible modifications of both renaming and truth-table are suggested.

Next, we give an scheme of the algorithm in a Pascal-like style. The notation used is the one of the previous sections.

```
Quasi-morphism_generator(($A_n$, T) ($B_m$, T'), f) =
        ; $A_n$ and $B_m$ are the truth-values sets,
        ; T and T' are the "and" connectives
        ; and f is a renaming from $A_n$ to $I(B_m)$
;1: Renaming checking
        if quasi-morphism?(A, B, f)
                ; A = ($A_n$, T) and B = ($B_m$, T')
                ; This predicate is implicitly defined
```

```
                    ; in the previous sections.
        Then Return (A, B, f)
                    ; If the renaming given by the expert
                    ; determines a quasi-morphism we return
                    ; it.
;2: Renaming generation
        R = {f: A_n → I(B_m) | quasi-morphism?(A, B, f) }
                    ; R is the set of all acceptable
                    ; renaming functions
                    ; The way they are computed is described
                    ; below.
        print(R)
        if select(r)
                    ; r ∈ R is selected by the expert
        then return (A, B, r)
;3: "And" truth-tables generation
        C = { T' / T' is an "and" operation on B_m and
        quasi-morphism?(A, (B_m, T'), f) }
        print(C)
        if select(c)
                    ; c ∈ C is selected by the expert
        then return (A, (B_m, c), f)
;4: Renaming and "and" truth-table generation
        RC = { (f, T') / f: A_n → I(B_m) and
        quasi-morphism?(A, (B_m, T'), f) }
        print(RC)
        if select((x, y))
                    ; (x,y) ∈ RC is selected by the expert
        then return (A, (B_m, y), x)
        else return nil
```

In the following a constructive description of predicates and functions used in the algorithm is given.

a) **Renaming checking**: It is only a checking predicate. It checks if a given function (renaming) is a quasi-morphism from A to B, where A and B are two given algebras.

b) **Renaming generation**: Let A and B be two truth-values algebras. *Renaming generation* is a predicate that returns all possible quasi-morphisms from the truth-values algebra A to the interval algebra I(B). The method to find these quasi-morphisms is the following:

  1. Define the operations T'* and N'* on I(B).

  2. Define mappings $f_1: N \cup F \longrightarrow N'^* \cup F'^*$ such that $f_1(0) = 0$, $f_1(F) = F'^*$ and $x \leq y$ implies $f_1(x) \flat f_1(y)$.

  3. For every $f_1$, define the negation morphism f by,

$$f(x) = \begin{cases} f_1(x), & \text{if } x \in N \cup F \\ N'^*(f_1(N(x))), & \text{if } x \in P \end{cases}$$

  4 Check if f is a quasi-morphism.

*Remarks*: The solutions, if any, contain as a particular case all the possible morphisms from A to B. By proposition 4.7, for every pair of algebras A and B, the *renaming generation* set (A,B) is not empty.

c) **"And" truth-table generation**: Let A be a truth-value algebra, let $B_m$ be a chain of m elements and let f be a renaming function from $A_n$ to $I(B_m)$. *"And" truth-table generation* is a predicate that returns "and" operations on $B_m$ such that f be a quasi-morphism. The method for finding "and" operations is the following:

  1. If f is not a negation morphism, it returns the empty set.
  2. Find all "and" operations T' on $B_m$.
  3. Check if f is a quasi-morphism.

*Remarks*: A general algorithm for finding truth-values algebras over a partially ordered set of n elements is given in [Godo and Meseguer, 1991].

d) **Renaming and truth-table generation**: Let A be a truth-value algebra and let $B_m$ be a chain of m elements. *Renaming and truth-table generation* is a function that, given $A_n$ and $B_m$, returns pairs (T',f), where T' is an "and" operation on $B_m$ and f is a quasi-morphism from A to <I($B_m$),T'*,N'*>. In this case, the method for finding pairs (T', f) is the following one:

  1. Define N'* on I($B_m$).
  2. Define all the possible functions $f_1$ as in the case of *renaming generation*.
  3. For every $f_1$, define the negation morphisms as in the case of *renaming generation*.
  4. Find all "and" operations T' on $B_m$.
  5. Check if f is a quasi-morphism.

*Final remarks:*
1.- All these methods can be automated.
2.- Some of the functions can return the empty set. (see examples of subsections 4.2 and 4.3)
3.- An order of preference on the sets *R*, *C* and *RC* should be given. There are many possibilities to do this. For example, several distances with respect to the initial renaming and truth-table operation can be defined on the above sets to obtain preference orderings. It seems reasonable that in all cases algebra morphisms should be preferred to algebra quasi-morphisms.
4.- The complexity of the algorithm is exponential in the number of truth-values. However, two factors make acceptable the execution time of the algorithm: (i) the number of different uncertainty linguistic terms used by an expert is usually not greater then nine [Miller, 1967], and (ii) the strong conditions required to "and" operations (see section 3) restrict the combinatorial explosion in the generation of such operations..

## 5 CONCLUSIONS AND FUTURE WORK

In this paper, the problem of communicating tasks with different uncertainty calculi has been introduced in a very general framework. This problem arises when dealing with large knowledge-based systems in which a better modelling of reasoning for different tasks requires working together with different uncertainty reasoning systems In the framework of rule-based expert systems, this problem has been analyzed in detail in the particular case in which different MV-logics are used to model the management of uncertainty in different tasks. Necessary and/or sufficient conditions for the correspondence mappings between them to be inference preserving have been given. Also an interactive algorithm to define such mappings has been proposed to assist the human expert in relating tasks or modules by inference preserving correspondences. However the general problem of communicating different uncertainty reasoning systems is very complex, and further research is needed as much in the field of distributed knowledge based systems as in the case of considering more complex models of uncertainty reasoning systems.

## Acknowledgements

This research has been partially supported by the CICYT id.880j382 project SPES, and by the ESPRIT-II Basic Research Action DRUMS.

## References

Agustí-Cullell J., Esteva F., Garcia P., Godo L.(1990) *'Formalizing Multiple-valuated Logics as Institutions'*.Short version in Proceedings of Third International IPMU Conference, Paris, pp.355-357. Full version will appear in Lecture Notes on Computer Science in June 1991.

Alsina C., Trillas E., Valverde L.(1983)' *On some logical connectives for Fuzzy Set Theory'*. Journal of Mathematical Analysis and Applications, 93, pp. 15-26.

Bonissone P.P. (1987) *'Summarizing and propagating uncertain information by triangular norms'* International Journal of aproximate reasoning 1 (1). pp. 71-101.

Godo L., Lopez de Mantaras R., Sierra C., Verdaguer A. (1989) *'MILORD:The architecture and management of linguistically expressed uncertainty'*. Int. Journal of Intelligent System Vol. 4 n° 4, pp. 471-501.

Godo L., Meseguer P. (1991) *'A constraint-based approach to generate finite truth-values algebras'*. Research Report IIIA-CEAB 91/9.

Goguen J., Burstall R.M., (1983) *'Introducing Institutions'* . Proc. Workshop on Logics of Programs, Carnegie-Mellon University. Springer LNCS 164, pp. 221-256.

Harper R., Sannella D., Tarleckil A. (1989) *'Structure and Representation in LF'* Proc.4th. IEEE Symposium on Logic of Computer Science.

Lopez de Mantaras R. (1990) *'Approximate Reasoning Models'* Elllis Horwood Ltd.

Miller G.A. (1967) *'The Magical Number Seven Plus or Minus Two: Some Limits on our Capacity for Processing Information'* in The Psychology of Communication, Penguin Books Inc.

Meseguer J. (1989) *'General Logics '*. In H.D. Ebbinghaus el al. (eds) Proc. Logic Colloquium '87. North - Holland.

Sierra C. (1989) ' *MILORD: Arquitectura multi-nivell per a sistemes experts en classificació*" Ph. D. Universitat Politècnica de Catalunya . Barcelona.

Trillas E., Valverde L. (1985) *'On mode and implication in approximate reasoning* ' in Gupta et al. (Eds.) Approximate reasoning in expert systems. pp. 157-166. North-Holland.

Verdaguer, A. (1989) *'PNEUMON-IA; desenvolupament i validació d'un sistema expert d'ajuda al diagnòstic mèdic'* Ph. D. Thesis. Universitat Autònoma de Barcelona.